\documentclass{article} 
\usepackage{iclr2016_workshop,times}
\usepackage{hyperref}
\usepackage{url}

\usepackage{subcaption}
\usepackage{graphicx}
\usepackage{amsmath}
\usepackage{amssymb}
\usepackage{amsfonts}
\renewcommand\bibsection{\subsection*{\refname}}
\def\bibfont{\small}
\newcommand{\Figref}[1]{Fig.~\ref{#1}}
\newcommand{\Eqref}[1]{Eq.~\ref{#1}}
\newcommand{\Eqrefs}[2]{Eqs.~\ref{#1},~\ref{#2}}

\title{Action Recognition using Visual Attention}

\author{%
Shikhar Sharma, Ryan Kiros \& Ruslan Salakhutdinov\\
Department of Computer Science\\
University of Toronto\\
Toronto, ON M5S 3G4, Canada\\
\texttt{\{shikhar,rkiros,rsalakhu\}@cs.toronto.edu}
}

\begin{document}

\maketitle

\begin{abstract}
We propose a soft attention based model for the task of action recognition in videos.
We use multi-layered Recurrent Neural Networks (RNNs) with Long Short-Term Memory (LSTM)
units which are deep both spatially and temporally. Our model learns to focus
selectively on parts of the video frames and classifies videos after taking a few
glimpses. The model essentially learns which parts in the frames are relevant for the
task at hand and attaches higher importance to them. We evaluate the model on UCF-11
(YouTube Action), HMDB-51 and Hollywood2 datasets and analyze how the model focuses its
attention depending on the scene and the action being performed.
\end{abstract}

\section{Introduction}
\vspace{-0.05in}
\label{sec:intro}
It has been noted in visual cognition literature that humans do not focus their
attention on an entire scene at once 
\citep{rensink2000dynamic}. Instead, they focus
sequentially on different parts of the scene to extract relevant information. Most
traditional computer vision algorithms do not employ attention mechanisms and are
indifferent to various parts of the image/video. With the recent surge of interest in deep neural networks, attention based models have
been shown to achieve promising results on several challenging tasks, including
caption generation \citep{DBLP:journals/corr/XuBKCCSZB15}, machine translation
\citep{DBLP:journals/corr/BahdanauCB14}, game-playing and tracking
\citep{DBLP:conf/nips/MnihHGK14}, as well as image recognition (e.g. Street View House
Numbers dataset \citep{DBLP:journals/corr/BaMK14}). Many of these models have employed
LSTM \citep{DBLP:journals/neco/HochreiterS97} based RNNs and have shown good results in
learning sequences. 

\begin{figure}[t]
    \begin{minipage}{\textwidth}
\hspace{0.15in}
    \begin{subfigure}{0.45\textwidth}
      \includegraphics[width=\linewidth]{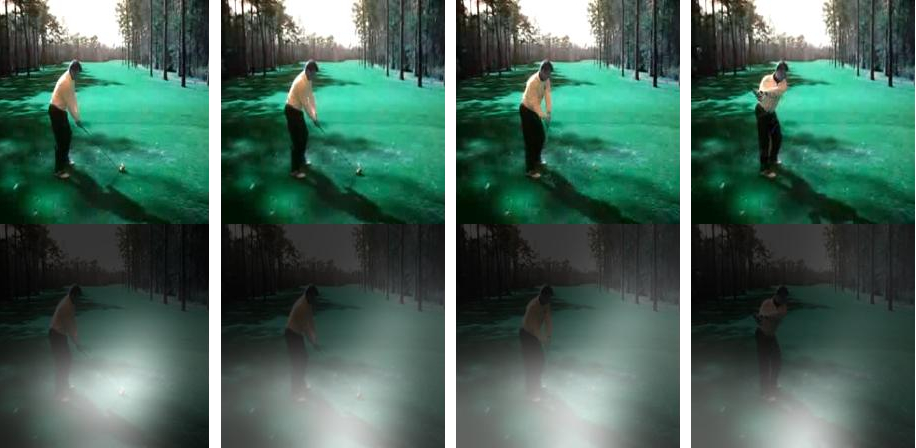}
      \caption{Correctly classified as ``golf swinging''}
      \label{subfig-intro-a}
    \end{subfigure}
\hspace{0.1in}
    \begin{subfigure}{0.45\textwidth}
      \includegraphics[width=\linewidth]{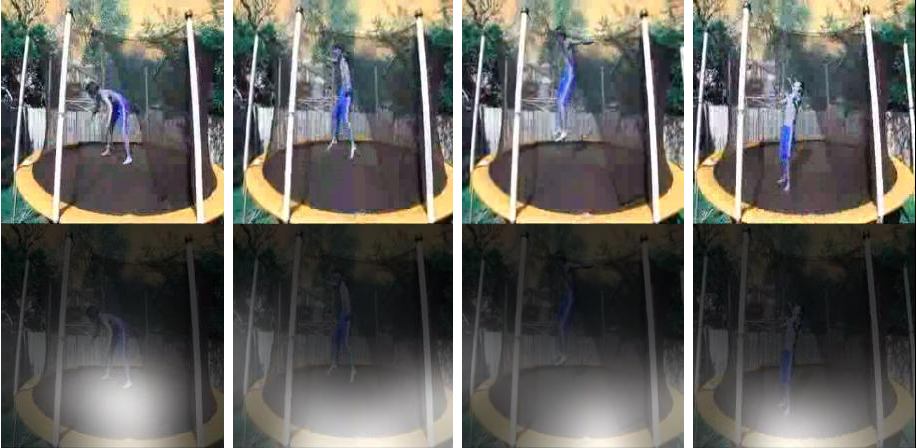}
      \caption{Correctly classified as ``trampoline jumping''}
      \label{subfig-intro-b}
    \end{subfigure}
    \end{minipage}
\vspace{-0.05in}
      \caption{\small
Attention over time:
The white regions show what the model is attending to and the
brightness indicates the strength of focus. Best viewed in color.} \label{fig:introfig}
\vspace{-0.2in}
\end{figure}

Attention models can be classified into soft attention and hard
attention models. Soft attention models are deterministic and can be trained using
backpropagation, whereas hard attention models are stochastic and can be trained by the
REINFORCE algorithm \citep{DBLP:journals/ml/Williams92,DBLP:conf/nips/MnihHGK14}, or by
maximizing a variational lower bound or using importance sampling~\citep{DBLP:journals/corr/BaMK14,Ba2015}. Learning hard
attention models can become computationally expensive as it requires sampling. In soft
attention approaches, on the other hand, a differentiable mapping can be used from all
the locations output to the next input.
Attention based models can also potentially infer the action
happening in videos by focusing only on the relevant places in each frame. For example,
\Figref{subfig-intro-a} shows four frames from the UCF-11 video sequence belonging
to the ``golf swinging'' category. The model tends to focus on the ball, the club, and
the human, which allows the model to correctly recognize the activity as
``golf swinging''.
In \Figref{subfig-intro-b},
our model attends to the trampoline, while correctly identifying the activity as
``trampoline jumping''.

In this paper we propose a soft attention based recurrent model for action recognition.
We describe how our model dynamically pools convolutional features and show that
using these features for action recognition gives better results compared to average or
max pooling which is used by many of the existing models~\citep{arXiv_video}. We further
demonstrate that our model tends to recognize important elements in video frames based 
on the activities it detects.

\section{Related Work}\label{sec:related}
\vspace{-0.05in}
Convolutional Neural Networks (CNNs) have been highly successful in image classification
and object recognition tasks \citep{DBLP:journals/corr/RenHGZ015,DBLP:journals/corr/WuYSDS15}. Classifying videos
instead of images adds a temporal dimension to the problem of image classification.
Learning temporal dynamics is a difficult problem and earlier approaches have used
optical flow, HOG and hand-crafted features to generate descriptors with both
appearance and dynamics information encoded. LSTMs have been recently shown to perform
well in the domain of speech recognition \citep{DBLP:conf/asru/GravesJM13},
machine translation \citep{Ilya}, image description 
\citep{DBLP:journals/corr/XuBKCCSZB15, DBLP:journals/corr/VinyalsTBE14} and video
description
\citep{DBLP:journals/corr/YaoTCBPLC15, DBLP:journals/corr/VenugopalanXDRMS14}.
They have also started picking up momentum in action recognition
\citep{DBLP:journals/corr/SrivastavaMS15,DBLP:journals/corr/NgHVVMT15}. 

Most of the existing approaches also tend to have CNNs underlying the LSTMs and classify
sequences directly or do temporal pooling of features prior to classification
\citep{DBLP:journals/corr/DonahueHGRVSD14,DBLP:journals/corr/NgHVVMT15}. LSTMs have also
been used to learn an effective representation of videos in unsupervised settings
\citep{DBLP:journals/corr/SrivastavaMS15} by using them in an encoder-decoder framework.
More recently, \cite{DBLP:journals/corr/YaoTCBPLC15} have proposed to use 3-D CNN
features and an LSTM decoder in an encoder-decoder framework to generate video
descriptions. Their model incorporates attention on a video level by defining a
probability distribution over frames used to generate individual words. They, however,
do not employ an attention mechanism on a frame level (i.e. within a single frame).

In general, it is rather difficult to
interpret internal representations learned by deep neural networks. Attention models add
a dimension of interpretability by capturing where the model is focusing its attention
when performing a particular task. 
\cite{DBLP:conf/cvpr/KarpathyTSLSF14} used a multi-resolution CNN
architecture to perform action recognition in videos. They mention the concept of fovea
but they fix attention to the center of the frame. A recent work of
\cite{DBLP:journals/corr/XuBKCCSZB15} used both soft attention and hard attention
mechanisms to generate image descriptions. Their model actually looks at the respective
objects when generating their description.
Our work directly builds upon this work.
However, while \cite{DBLP:journals/corr/XuBKCCSZB15} primarily worked on caption
generation in static images, in this paper, we focus on using a soft attention mechanism
for activity recognition in videos.
More recently, 
\cite{DBLP:journals/corr/JaderbergSZK15} have 
proposed a soft-attention mechanism called the \textit{Spatial Transformer} module which 
they add between the layers of CNNs. Instead of weighting locations using a softmax 
layer which we do, they apply affine transformations to multiple layers of their CNN 
to attend to the relevant part and get state-of-the-art results on the Street View House Numbers dataset \citep{svhn}.
\cite{DBLP:journals/corr/YeungRJAML15} do dense action labelling using a
temporal attention based model on the input-output context and report higher
accuracy and better understanding of temporal relationships in action videos.

\section{The Model and the Attention Mechanism}\label{sec:attmodel}
\vspace{-0.05in}
\subsection{Convolutional Features}
\vspace{-0.05in}
We extract the last convolutional layer obtained by pushing the video frames through
GoogLeNet model~\citep{DBLP:journals/corr/SzegedyLJSRAEVR14} trained on the ImageNet
dataset~\citep{DBLP:conf/cvpr/DengDSLL009}. This last convolutional layer has $D$
convolutional maps and is a feature cube of shape $K \times K \times D$ ($7 \times 7 \times 1024$ in our experiments). Thus, at each
time-step $t$, we extract $K^2$ $D$-dimensional vectors. We refer to these vectors as
feature slices in a feature cube:
\begin{align*}
\mathbf{X}_t &=  [\mathbf{X}_{t,1}, \dots, \mathbf{X}_{t,K^2}], & \mathbf{X}_{t,i} \in \mathbb{R}^D.
\end{align*}
Each of these $K^2$ vertical feature slices maps to different overlapping regions in the
input space and our model chooses to focus its attention on these $K^2$ regions.

\subsection{The LSTM and the Attention Mechanism}
\vspace{-0.05in}
We use the LSTM implementation discussed in \cite{DBLP:journals/corr/ZarembaSV14} and
\cite{DBLP:journals/corr/XuBKCCSZB15}:
\begin{align}
\begin{pmatrix}\label{eq:1}
        \mathbf{i}_t\\
        \mathbf{f}_t\\
        \mathbf{o}_t\\
        \mathbf{g}_t
        \end{pmatrix} &= \begin{pmatrix}
                                         \sigma\\
                                         \sigma\\
                                         \sigma\\
                                         \text{tanh}
                         \end{pmatrix} M \begin{pmatrix}
                                                  \mathbf{h}_{t-1},\\
                                                  \mathbf{x}_t
                                         \end{pmatrix},\\
\mathbf{c}_t &= \mathbf{f}_t \odot \mathbf{c}_{t-1} + \mathbf{i}_t \odot \mathbf{g}_t,\\
\mathbf{h}_t &= \mathbf{o}_t \odot \text{tanh}(\mathbf{c}_t),
\end{align}

where $\mathbf{i}_t$ is the input gate, $\mathbf{f}_t$ is the forget gate,
$\mathbf{o}_t$ is the output gate, and $\mathbf{g}_t$ is calculated as shown 
in~\Eqref{eq:1}. $\mathbf{c}_t$ is the cell state, $\mathbf{h}_t$ is the hidden
state, and $\mathbf{x}_t$ (see \Eqrefs{eq:4}{eq:5}) represents the input
to the LSTM at time-step $t$. $M: \mathbb{R}^a \to \mathbb{R}^b$ is an affine
transformation consisting of trainable parameters with $a=d+D$ and $b=4d$, where $d$ is
the dimensionality of all of $\mathbf{i}_t$, $\mathbf{f}_t$, $\mathbf{o}_t$,
$\mathbf{g}_t$, $\mathbf{c}_t$, and~$\mathbf{h}_t$.

\begin{figure}[!t]
\vspace{-0.1in}
    \begin{minipage}{\textwidth}
    \begin{subfigure}{0.37\textwidth}
      \includegraphics[width=0.9\linewidth]{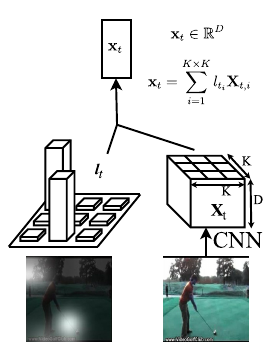}
      \caption{The soft attention mechanism}
      \label{fig:attention-mechanism}
    \end{subfigure}%
    \begin{subfigure}{0.06\textwidth}\hfill
    \end{subfigure}%
    \begin{subfigure}{0.5\textwidth}
      \includegraphics[width=\linewidth]{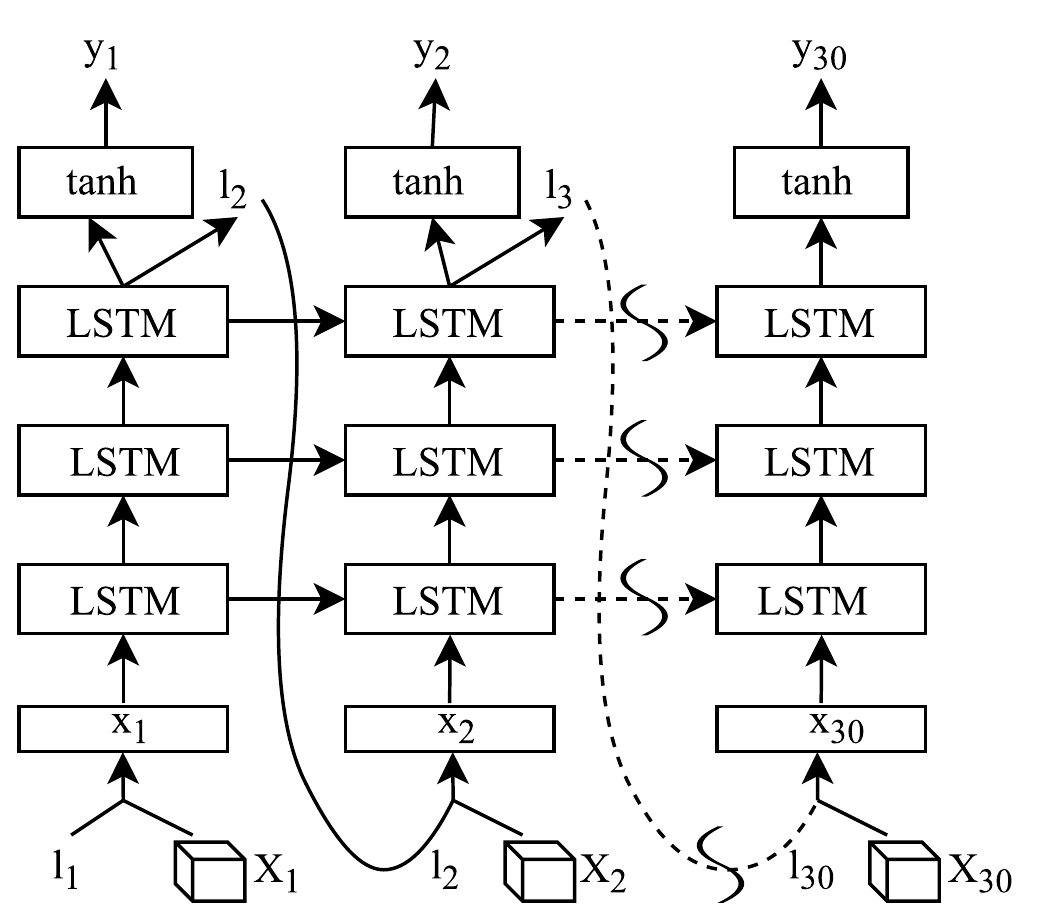}
      \caption{Our recurrent model}
      \label{fig:model}
    \end{subfigure}
    \end{minipage}
\vspace{-0.05in}
      \caption{(\ref{fig:attention-mechanism}) \small The CNN takes the video frame as
its input and produces a feature cube. The model computes the current input $\textbf{x}_t$
as an average of the feature slices weighted according to the location softmax $\textbf{l}_{t}$
(\ref{fig:model}) At each time-step $t$, our recurrent network takes a feature slice
$\textbf{x}_t$, generated as in (\ref{fig:attention-mechanism}), as the input. It then propagates
$\textbf{x}_t$ through three layers of LSTMs and predicts the next location probabilities
$\textbf{l}_{t+1}$ and the class label~$\textbf{y}_t$.}
    \vspace{-0.15in}
\end{figure}

At each time-step $t$, our model predicts $\mathbf{l}_{t+1}$, a softmax over $K \times
K$ locations, and $\mathbf{y}_t$, a softmax over the label classes with an additional
hidden layer with $tanh$ activations (see \Figref{fig:model}). The location softmax
is defined as follows:
\vspace{-1mm}
\begin{align}
l_{t,i}=p(\mathbf{L}_{t}=i|\mathbf{h}_{t-1}) &= \frac{\exp(W_i^\top \mathbf{h}_{t-1})}{\sum_{j=1}^{K \times K} \exp(W_j^\top \mathbf{h}_{t-1})}\qquad i\in 1\dots K^2, \label{eq:4}
\end{align}
where $W_i$ are the weights mapping to the $i^{th}$ element of the location softmax and
$\mathbf{L}_t$ is a random variable which can take 1-of-$K^2$ values. This softmax can
be thought of as the probability with which our model believes the corresponding region
in the input frame is important. After calculating these probabilities, the soft
attention mechanism~\citep{DBLP:journals/corr/BahdanauCB14}
computes the expected value of the input
at the next time-step $\mathbf{x}_t$ by taking expectation over the feature slices at
different regions (see \Figref{fig:attention-mechanism}):
\vspace{-1mm}
\begin{align}
\mathbf{x}_t = \mathbb{E}_{p(\mathbf{L}_t|\mathbf{h}_{t-1})}[\mathbf{X}_t] &= \sum_{i=1}^{K^2} l_{t,i} \mathbf{X}_{t,i}, \label{eq:5}
\end{align}
where $\mathbf{X}_t$ is the feature cube and $\mathbf{X}_{t,i}$ is the $i^{th}$ slice of
the feature cube at time-step $t$. Note that in the hard attention based models, we
would sample $\mathbf{L}_t$ from a softmax distribution of \Eqref{eq:4}. The input
$\mathbf{x}_t$ would then be the feature slice at the sampled location instead of
taking expectation over all the slices. Thus, hard attention based models are not
differentiable and have to resort to some form of sampling.

We use the following initialization strategy (see \cite{DBLP:journals/corr/XuBKCCSZB15}) for the cell state and the hidden state of the LSTM for faster convergence:
\begin{align}
\mathbf{c}_0 = f_{\text{init,c}}\left(\frac{1}{T}\sum_{t=1}^T \left(\frac{1}{K^2}\sum_{i=1}^{K^2} \mathbf{X}_{t,i}\right)\right)
\quad \text{and} \quad
\mathbf{h}_0 = f_{\text{init,h}}\left(\frac{1}{T}\sum_{t=1}^T\left(\frac{1}{K^2}\sum_{i=1}^{K^2} \mathbf{X}_{t,i}\right) \right),\label{eq:3}
\end{align}
where $f_{\text{init,c}}$ and $f_{\text{init,h}}$ are two multilayer perceptrons and $T$
is the number of time-steps in the model. These values are used to calculate the first
location softmax $\mathbf{l}_1$ which determines the initial input $\mathbf{x}_1$. In
our experiments, we use multi-layered deep LSTMs, as shown in \Figref{fig:model}.

\subsection{Loss Function and the Attention Penalty}
We use cross-entropy loss coupled with the doubly stochastic penalty introduced in 
\cite{DBLP:journals/corr/XuBKCCSZB15}. We impose an additional constraint over the
location softmax, so that $\sum_{t=1}^T l_{t,i} \approx 1$. This is the attention
regularization which forces the model to look at each region of the frame at some point
in time. The loss function is defined as follows:
\vspace{-0.05in}
\begin{align}
L &= -\sum_{t=1}^T\sum_{i=1}^{C} y_{t,i} \log \hat{y}_{t,i} + \lambda\sum_{i=1}^{K^2}(1-\sum_{t=1}^T l_{t,i})^2 + \gamma \sum_{i}\sum_{j}\theta_{i,j}^2,
\label{eq:loss}
\end{align}
\vspace{-0.1in}
\\where $y_t$ is the one hot label vector,
$\hat{y}_t$ is the vector of class probabilities at time-step $t$, $T$ is the total
number of time-steps, $C$ is the number of output classes, $\lambda$ is the attention
penalty coefficient, $\gamma$ is the weight decay coefficient, and $\theta$ represents
all the model parameters. 
Details about the architecture and hyper-parameters
are given in Section~\ref{subsec:evaluation}.

\section{Experiments}\label{sec:experiments}
\vspace{-0.05in}
\subsection{Datasets}
\vspace{-0.05in}
We have used UCF-11, HMDB-51 and Hollywood2 datasets in our experiments. UCF-11 is the
YouTube Action dataset consisting of 1600 videos and 11 actions - basketball shooting,
biking/cycling, diving, golf swinging, horse back riding, soccer juggling, swinging,
tennis swinging, trampoline jumping, volleyball spiking, and walking with a dog. The
clips have a frame rate of 29.97 fps and 
each video has only one action associated with it. We use 975 videos for training and 625
videos for testing.

HMDB-51 Human Motion Database dataset provides three train-test splits each consisting
of 5100 videos. These clips are labeled with 51 classes of human actions like Clap,
Drink, Hug, Jump, Somersault, Throw and many others. Each video has only one action
associated with it. The training set for each split has 3570 videos (70 per category)
and the test set has 1530 videos (30 per category). The clips have a frame rate of 30
fps.

Hollywood2 Human Actions dataset consists of 1707 video clips collected from movies.
These clips are labeled with 12 classes of human actions - AnswerPhone, DriveCar, Eat,
FightPerson, GetOutCar, HandShake, HugPerson, Kiss, Run, SitUp, SitDown and StandUp.
Some videos have multiple actions associated with them. The training set has 823 videos
and the testing set has 884 videos.

All the videos in the datasets were resized to $224 \times 224$ resolution and fed to a
GoogLeNet model trained on the ImageNet dataset. The last convolutional layer of size
$7 \times 7 \times 1024$ was used as an input to our model.

\begin{table}[t]
\caption{Performance on UCF-11 (acc \%), HMDB-51 (acc \%) and Hollywood2 (mAP \%)}
\vspace{-0.1in}
\label{table:ucf11holly2}
\begin{center}
\begin{tabular}{l||c||c||c}
\multicolumn{1}{c||}{\bf Model}  &\multicolumn{1}{c||}{\bf UCF-11} &\multicolumn{1}{c||}{\bf HMDB-51} &\multicolumn{1}{c}{\bf Hollywood2}
\\ \hline\hline
Softmax Regression (full CNN feature cube)    & 82.37 & 33.46 & 34.62 \\
\hline
Avg pooled LSTM (@ 30 fps)                    & 82.56 & 40.52 & 43.19 \\
Max pooled LSTM (@ 30 fps)                    & 81.60 & 37.58 & 43.22 \\
\hline\hline
Soft attention model (@ 30 fps, $\lambda=0$)  & 84.96 & 41.31 & 43.91 \\
Soft attention model (@ 30 fps, $\lambda=1$)  & 83.52 & 40.98 & 43.18 \\
Soft attention model (@ 30 fps, $\lambda=10$) & 81.44 & 39.87 & 42.92
\end{tabular}
\end{center}
\vspace{-0.05in}
\end{table}

\begin{figure}[t]
    \begin{subfigure}{\textwidth}
      \includegraphics[width=\linewidth]{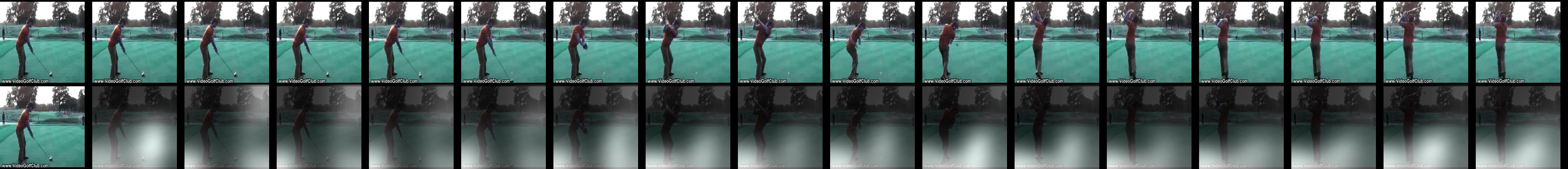}
      \caption{$\lambda = 0$}
      \label{subfig-alpha-a}
    \end{subfigure}
    \begin{subfigure}{\textwidth}
      \includegraphics[width=\linewidth]{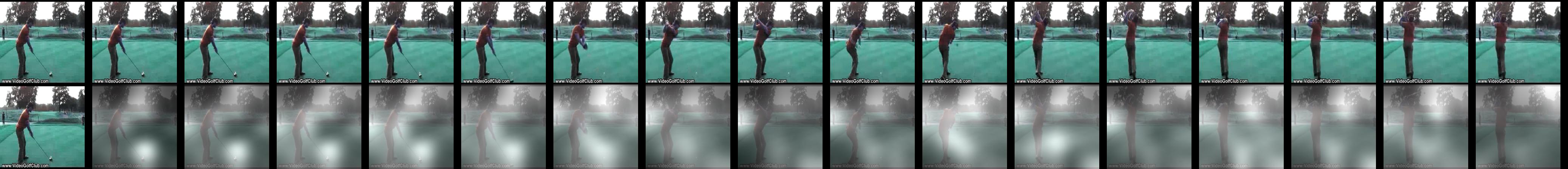}
      \caption{$\lambda = 1$}
      \label{subfig-alpha-b}
    \end{subfigure}
    \begin{subfigure}{\textwidth}
      \includegraphics[width=\linewidth]{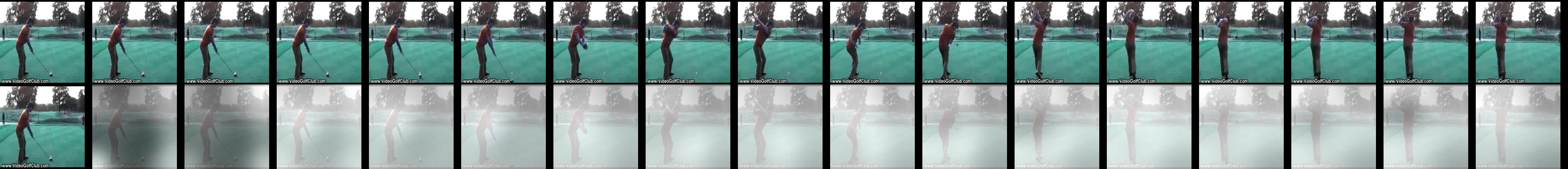}
      \caption{$\lambda = 10$}
      \label{subfig-alpha-c}
    \end{subfigure}
\vspace{-0.1in}
      \caption{\small
Variation in the model's attention depending on the value of attention penalty
$\lambda$. The white regions are where the model is looking and the brightness indicates
the strength of focus. Setting $\lambda=0$ corresponds to the model that tends to select
a few locations and stay fixed on them. Setting $\lambda=10$ forces the model to gaze
everywhere, which resembles average pooling over slices.
}\label{fig:penalties}
     \vspace{-0.2in}
\end{figure}

\begin{table}[t]
\vspace{-0.3in}
\caption{Comparison of performance on HMDB-51 and Hollywood2 with state-of-the-art
models}
\vspace{-0.1in}
\label{table:hmdb51comparison}
\begin{center}
\begin{tabular}{l||c||c}
\multicolumn{1}{c||}{\bf Model} &\multicolumn{1}{c||}{\bf HMDB-51} &\multicolumn{1}{c}{\bf Hollywood2}\\
\multicolumn{1}{c||}{} &\multicolumn{1}{c||}{\bf (acc \%)} &\multicolumn{1}{c}{\bf (mAP \%)}
\\ \hline\hline
Spatial stream ConvNet \hfill     \citep{DBLP:journals/corr/SimonyanZ14} & 40.5 & -    \\
Soft attention model \hfill                                  (Our model) & 41.3 & 43.9 \\
Composite LSTM Model \hfill    \citep{DBLP:journals/corr/SrivastavaMS15} & 44.0 & -    \\
DL-SFA \hfill                          \citep{DBLP:conf/cvpr/SunJCFWY14} & -    & 48.1 \\
\hline
Two-stream ConvNet \hfill         \citep{DBLP:journals/corr/SimonyanZ14} & 59.4 & -    \\
VideoDarwin        \hfill                          \citep{Fernando2015a} & 63.7 & 73.7 \\
Multi-skIp Feature Stacking \hfill  \citep{DBLP:journals/corr/LanLLHR14} & 65.1 & 68.0 \\
Traditional+Stacked Fisher Vectors\hfill\citep{DBLP:conf/eccv/PengZQP14} & 66.8 & -    \\
\hline
Objects+Traditional+Stacked Fisher Vectors \hfill     \citep{JainCVPR15} & 71.3 & 66.4
\end{tabular}
\end{center}
\vspace{-0.00in}
\end{table}
\begin{figure}[t]
     \vspace{-0.0in}
    \begin{minipage}{\textwidth}
\hspace{0.4in}
    \begin{subfigure}{0.43\textwidth}
      \includegraphics[width=\linewidth]{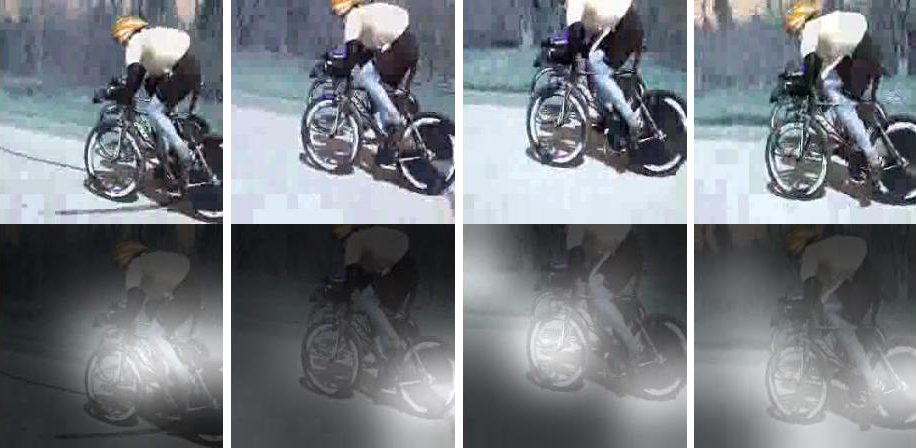}
      \caption{Correctly classified as ``cycling''}
      \label{subfig-mid-a}
    \end{subfigure}
\hspace{0.1in}
    \begin{subfigure}{0.43\textwidth}
      \includegraphics[width=\linewidth]{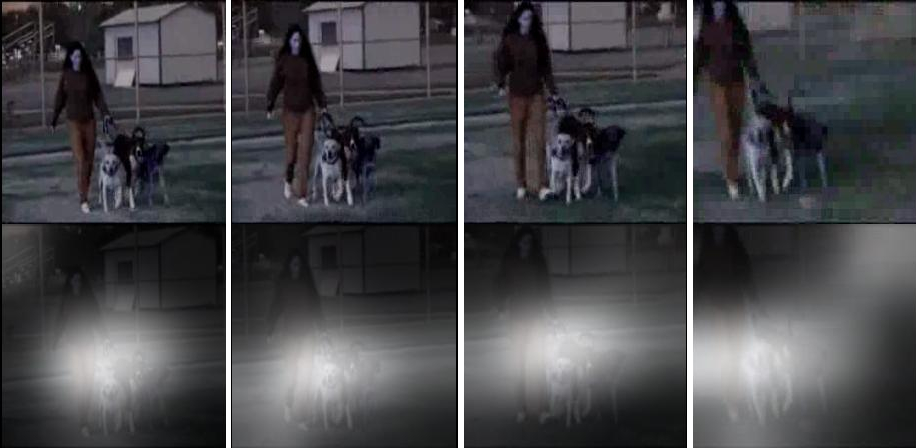}
      \caption{ Correctly classified as ``walking with a dog''}
      \label{subfig-mid-b}
    \end{subfigure}
    \end{minipage}
  \vspace{-0.1in}
      \caption{\small Attention over time.
The model learns to look at
the relevant parts - the cycle frame in (a) and the human and the dogs in (b)}
\label{fig:midfig}
    \vspace{-0.0in}
\end{figure}

\begin{figure}[t!]
    \begin{minipage}{0.43\textwidth}
\hspace{0.4in}
    \begin{subfigure}{\textwidth}
      \includegraphics[width=\linewidth]{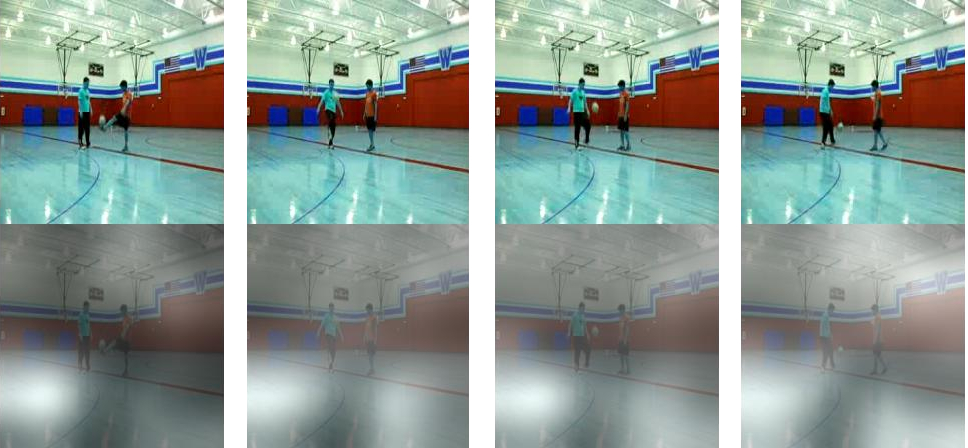}
      \caption{Incorrectly classified as ``diving''}
      \label{subfig-c}
    \end{subfigure}
    \end{minipage}%
    \begin{minipage}{0.04\textwidth}
    \end{minipage}%
    \hspace{0.6in}
    \begin{minipage}{0.37\textwidth}
    \begin{subfigure}{\textwidth}
      \includegraphics[width=\linewidth]{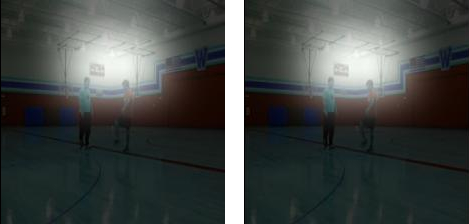}
      \caption{Incorrectly classified as ``volleyball''}
      \label{subfig-d}
    \end{subfigure}
    \end{minipage}
  \vspace{-0.1in}
    \caption{\small Video frames for a few time-steps for an example of soccer played on
a basketball court. 
Different glimpses can result in different
predictions. Best viewed in color.} \label{fig:attention}
    \vspace{-0.10in}
\end{figure}

\subsection{Training Details and Evaluation}\label{subsec:evaluation}
\vspace{-0.05in}
In all of our experiments, model architecture and various other hyper-parameters
were set using cross-validation.
In particular, for all datasets we trained 3-layer LSTM models, where
the dimensionality of the LSTM hidden state, cell state, and
the hidden layer were set to 512 for both UCF-11 and Hollywood2 and 1024 for HMDB-51.
We also experimented with models having one LSTM layer to five LSTM layers, but
did not observe any significant improvements in model performance.
For the attention penalty coefficient we experimented with values 0,
1, 10. While reporting results, we have set the weight decay penalty to $10^{-5}$ and
use dropout \citep{DBLP:journals/jmlr/SrivastavaHKSS14} of 0.5 at all non-recurrent
connections.
All models were trained using Adam optimization algorithm~\citep{DBLP:journals/corr/KingmaB14}
for 15 epochs over the entire datasets. However, we found that
Adam usually converged after 3 epochs.
Our implementation is based in Theano~\citep{theano} which also
handles the gradient computation and our code is available at \url{https://github.com/kracwarlock/action-recognition-visual-attention}.

For both training and testing our model takes 30 frames at a time sampled at fixed $fps$
rates. We split each video into groups of 30 frames starting with the first frame,
selecting 30 frames according to the $fps$ rate, and then moving ahead with a stride of
1. Each video thus gets split into multiple 30-length samples.
At test time, we compute class predictions for each time step and then average
those predictions over 30 frames.
To obtain a prediction for the entire video clip,
we average the predictions from all 30 frame blocks in the video.

\subsubsection{Baselines}
\vspace{-0.05in}
The softmax regression model uses the complete $7 \times 7 \times 1024$ feature cube as
its input to predict the label at each time-step $t$, while all other models use only a
$1024$-dimensional feature slice as their input. The average pooled and max pooled LSTM
models use the same architecture as our model except that they do not have any
attention mechanism and thus do not produce a location softmax. The inputs at each
time-step for these models are obtained by doing average or max pooling over the $7
\times 7 \times 1024$ cube to get 1024 dimensional slices, whereas our soft attention
model dynamically weights the slices by the location softmax (see \Eqref{eq:5}).

\subsection{Quantitative analysis}
\vspace{-0.05in}
Table~\ref{table:ucf11holly2} reports accuracies on both UCF-11 and HMDB-51 datasets and
mean average precision (mAP) on Hollywood2. 
Even though the softmax regression baseline is given the
complete $7 \times 7 \times 1024$ cube as its input,
it performs worse than our
model for all three datasets and worse than all models in the case of HMDB-51 and
Hollywood2. The results from Table~\ref{table:ucf11holly2} demonstrate that our
attention model performs better than both average and max pooled LSTMs.

We next experimented with doubly stochastic penalty term $\lambda$ (see \Eqref{eq:loss}). 
Figure \ref{subfig-alpha-a} shows that with no attention regularization term, $\lambda = 0$,
the model tends to vary its attention less. 
Setting $\lambda = 1$ encourages the model to further explore different gaze locations.  
The model with
$\lambda = 10$ looks everywhere (see \Figref{subfig-alpha-c}), in which case its
behavior tends to become similar to the average pooling case. Values in between these
correspond to dynamic weighted averaging of the slices. The models with $\lambda=0$ and
$\lambda = 1$ perform better than the models with $\lambda = 10$.

In Table~\ref{table:hmdb51comparison}, we compare the performance of our model with
other state-of-the-art action recognition models.
We do not include UCF-11 here due to the lack of standard train-test splits.
We have divided the table into three
sections. Models in the first section use only RGB data while models in the second
section use both RGB and optical flow data. The model in the third section uses both RGB,
optical 
flow, as well as object responses of the videos on some ImageNet categories.
Our model performs competitively against deep learning models in its category
(models using RGB features only), while providing some insight 
into where the neural network is looking.

\subsection{Qualitative analysis}
\vspace{-0.05in}
Figure~\ref{fig:midfig} shows some test examples of where our model attends to on UCF-11
dataset. In \Figref{subfig-mid-a}, we see that the model was able to focus on parts
of the cycle, while correctly recognizing the activity as ``cycling''. Similarly, in
\Figref{subfig-mid-b}, the model attends to the dogs and classifies the activity as
``walking with a dog''.

\begin{figure}[t]
\vspace{-0.2in}
  \begin{minipage}{\textwidth}
\hspace{0.4in}
    \begin{subfigure}{0.41\textwidth}
      \includegraphics[width=\linewidth]{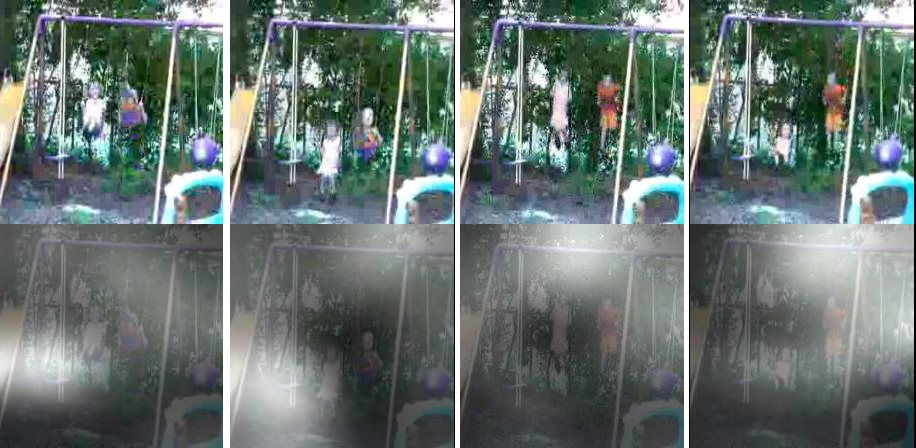}
      \caption{Correctly classified as ``swinging''}
      \label{subfig-ev-a}
    \end{subfigure}
\hspace{0.1in}
    \begin{subfigure}{0.41\textwidth}
      \includegraphics[width=\linewidth]{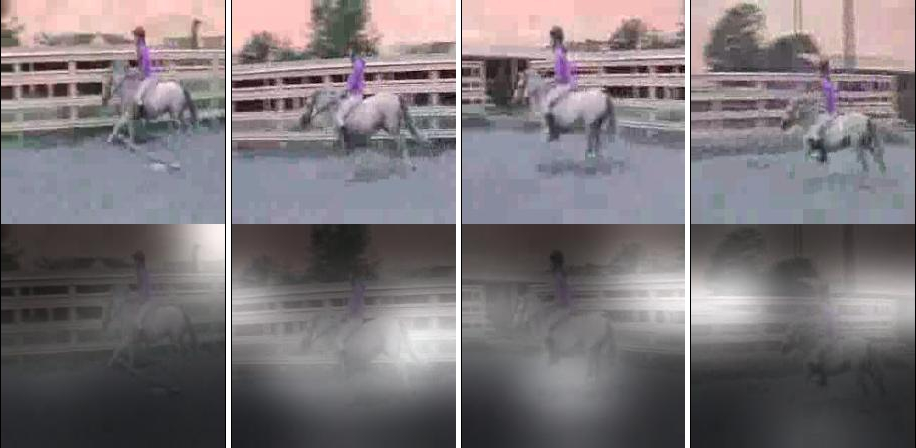}
      \caption{Correctly classified as ``horse back riding''}
      \label{subfig-ev-b}
    \end{subfigure}
  \end{minipage}
    \vspace*{-0.1in}
      \caption{\small Video frames where the model pays more attention to the background
compared to the foreground and still classifies them correctly} \label{fig:everywhere}
  \vspace{-0.05in}
\end{figure}
\begin{figure}[t]
    \begin{minipage}{\textwidth}
    \hspace{0.4in}
    \begin{subfigure}{0.41\textwidth}
      \includegraphics[width=\linewidth]{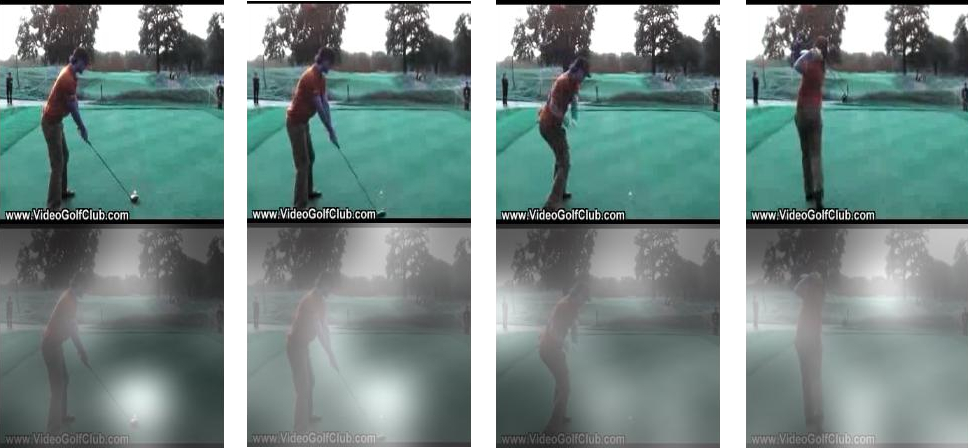}
      \caption{``golf swinging'' (@ 6fps, $\lambda=1$)}
      \label{subfig-6fps}
    \end{subfigure}
     \hspace{0.1in}
    \begin{subfigure}{0.41\textwidth}
      \includegraphics[width=\linewidth]{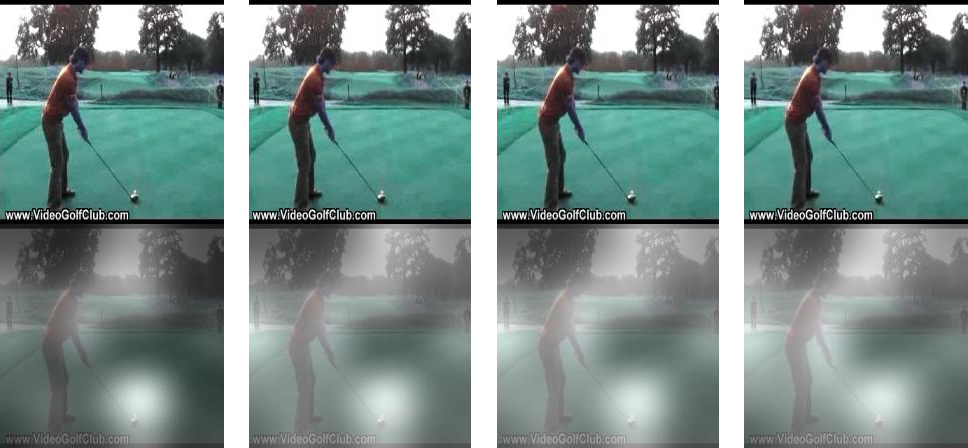}
      \caption{``golf swinging'' (@ 30 fps, $\lambda=1$)}
      \label{subfig-30fps}
    \end{subfigure}
    \end{minipage}
    \vspace*{-0.1in}
      \caption{\small The model's focus of attention visualized over four equally spaced
timesteps at different fps rates. (a) plays faster and when the ball is hit and the club
disappears, the model searches around to find them. (b) plays slower and the model stays
focused on the ball and the club.} \label{fig:fpsfig}
    \vspace*{-0.0in}
\end{figure}
\begin{figure}[t!]
    \vspace*{-0.05in}
    \begin{minipage}{\textwidth}
    \hspace{0.4in}
    \begin{subfigure}{0.41\textwidth}
      \includegraphics[width=\linewidth]{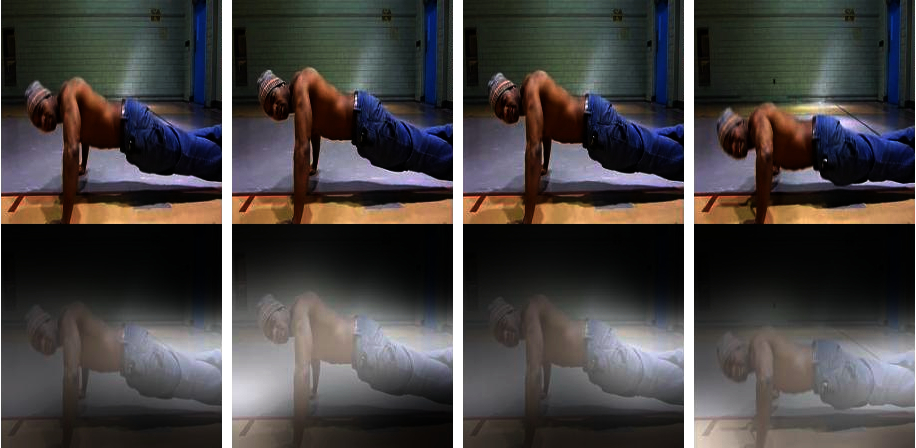}
      \caption{Correctly classified as ``Pushup''}
      \label{pushup}
    \end{subfigure}
    \hspace{0.1in}
    \begin{subfigure}{0.41\textwidth}
      \includegraphics[width=\linewidth]{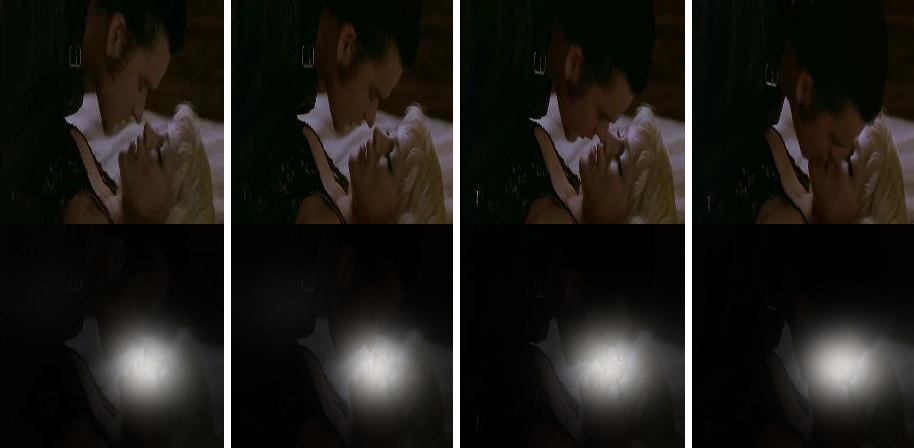}
      \caption{Correctly classified as ``Kiss''}
      \label{kiss-h2}
    \end{subfigure}
     \vspace{0.1in}
    \end{minipage}
    \begin{minipage}{\textwidth}
    \hspace{0.4in}
    \begin{subfigure}{0.41\textwidth}
      \includegraphics[width=\linewidth]{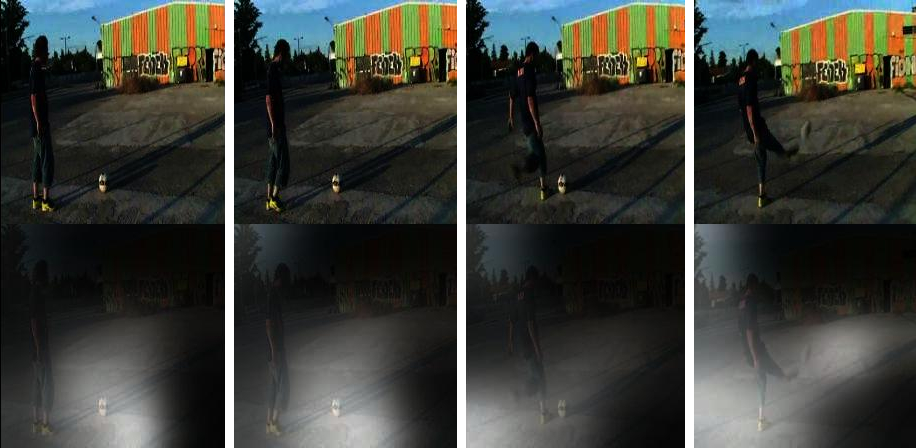}
      \caption{Inorrectly classified as ``Somersault''}
      \label{kick_ball}
    \end{subfigure}
    \hspace{0.1in}
    \begin{subfigure}{0.41\textwidth}
      \includegraphics[width=\linewidth]{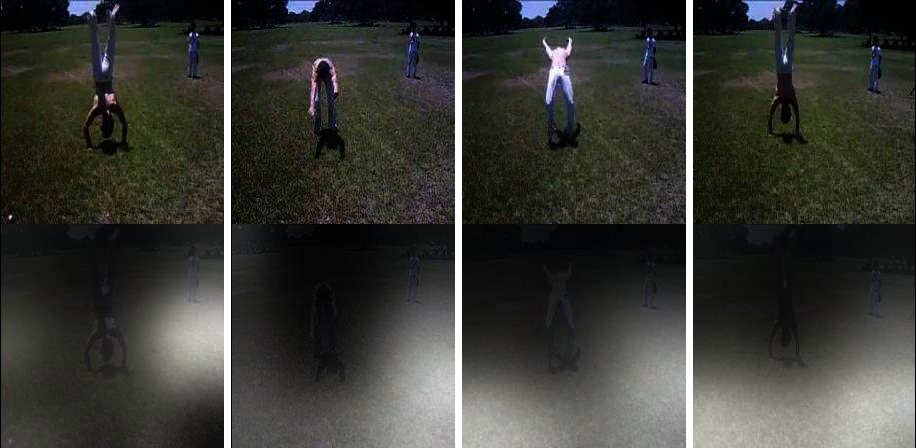}
      \caption{Incorrectly classified as ``Hit''}
      \label{flic_flac}
    \end{subfigure}
    \end{minipage}
    \vspace*{-0.1in}
      \caption{\small Visualization of the focus of attention for four videos from
HMDB-51 and Hollywood2 datasets over time. The white regions are where the model is looking and the
brightness indicates the strength of focus.} \label{fig:atth2}
    \vspace*{-0.15in}
\end{figure}
We can also better understand failures of the model using the attention mechanism. For
example, \Figref{subfig-c} shows that the model mostly attends to the background
like the light blue floor of the court. The model incorrectly classifies the example as ``diving''. However, using a different
manually specified glimpse, as shown in \Figref{subfig-d}, the model classifies the same example as
``volleyball spiking''. It is quite interesting to see that we can better understand
the success and failure cases of this deep attention model by visualizing where it
attends to.\footnote{All the figures are from 
our best performing models with $\lambda=0$ unless otherwise mentioned.}

The model does not always need to attend to the foreground. In many cases the camera is
far away and it may be difficult to make out what the humans are doing or what the
objects in the frames are. In these cases the model tends to look at the background and
tries to infer the activity from the information in the background. For example, the
model can look at the basketball court in the background and predict the action being
performed. Thus, depending on the video both
foreground and background might be important for activity recognition. Some 
examples are shown in \Figref{fig:everywhere}, where the model appears to look
everywhere.

\begin{figure}[t!]
\vspace{-0.2in}
\begin{minipage}{0.65\textwidth}
\includegraphics[width=0.95\textwidth]{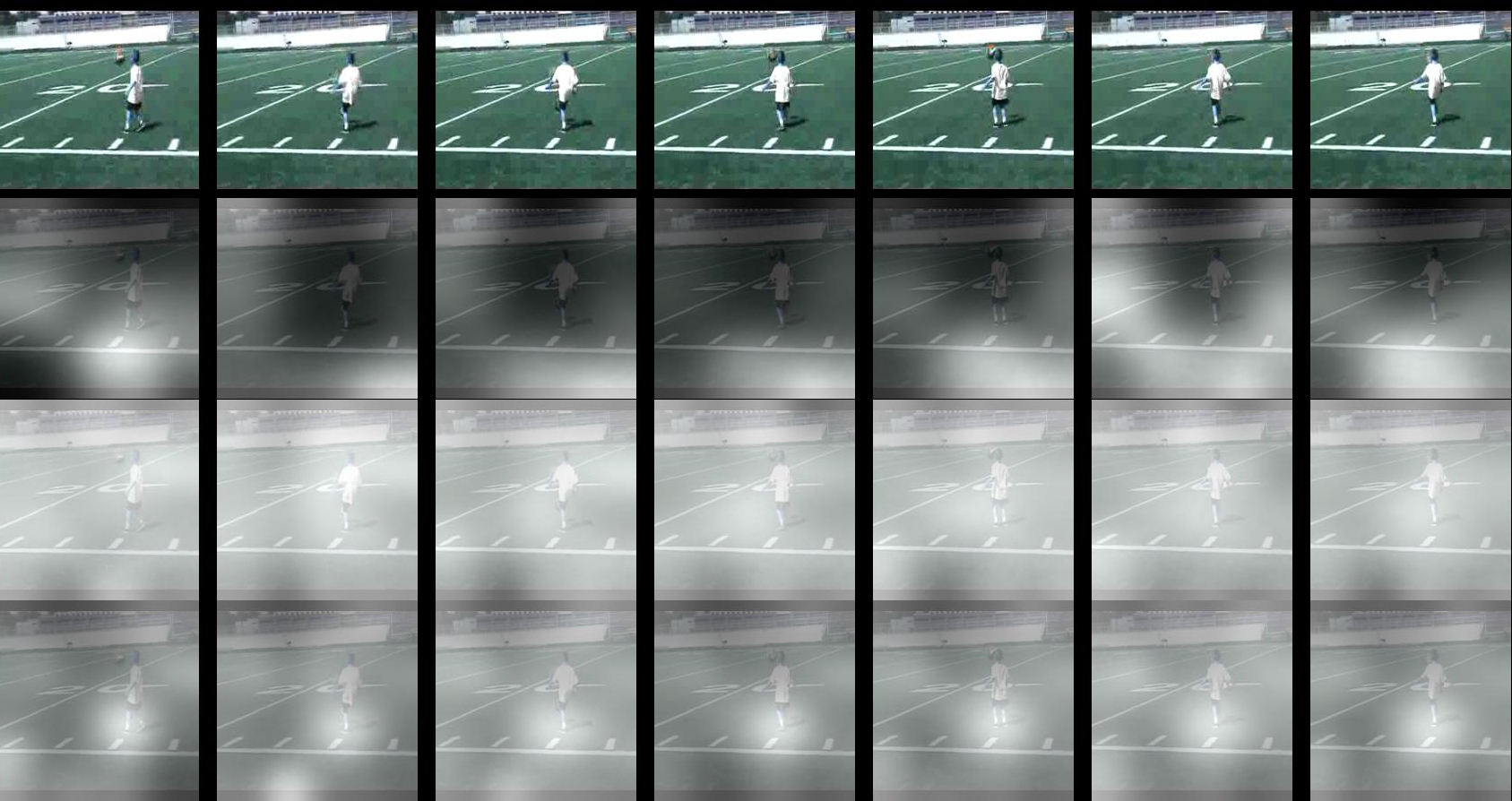}
\end{minipage}%
\begin{minipage}{0.35\textwidth}
\caption{\small (\textbf{First}) The original video frames for a ``soccer juggling''
example from UCF-11 (\textbf{Second}) Glimpse of model with $\lambda = 1$ overlayed
on the frames; predicted incorrectly as ``tennis swinging'' (\textbf{Third}) Randomly
initialized glimpse overlayed on the frames; predicted incorrectly as
``tennis swinging'' (\textbf{Fourth}) The first glimpse at which the action is
correctly predicted as ``soccer juggling", overlayed on the frames}
\label{fig:blankagain}
\end{minipage}
\vspace{-0.15in}
\end{figure}

It is also interesting to observe that in some cases, the model is able to attend to
important objects in the video frames and attempts to track them to some extent in order
to correctly identify the performed activity. In \Figref{subfig-30fps}, the video is
sampled at 30fps and subsequent frames are almost identical. In this case the model
stays focused on the golf ball, club, and the human. However, when we change the
sampling rate to 6fps, as shown in \Figref{subfig-6fps}, we find that the video
frames change quickly. The model now remains focused on the ball before it disappears.
After the person hits the ball, we see that the model tries to look at other places,
possibly to track the ball and the golf club.

We next examined the model's performance on the HMDB-51 dataset.\footnote{More 
examples of our model's attention are available in Appendix \ref{app:examples} and  
at\\ \url{http://www.cs.toronto.edu/~shikhar/projects/action-recognition-attention}.}
In \Figref{pushup} the model attempts to focus
on the person performing push-ups to recognize ``Pushup'' activity.
In \Figref{kick_ball} the model classifies the example of 
``KickBall'' incorrectly as ``Somersault'' despite attending to the location where the 
action is happening. In some cases, however, the model fails to even attend 
to the relevant location (see \Figref{flic_flac}).
For Hollywood2, \Figref{kiss-h2} shows an example of a short clip belonging to the
``Kiss'' action. It appears that the model correctly anticipates that a kiss is going to
take place and attempts to focus on the region between the man and the woman.

In our final set of experiments, we have tried to examine 
some failure cases of our attention mechanism. 
As an example, \Figref{fig:blankagain} 
shows a test video clip of ``soccer juggling'' (top row). Our
model focuses on the white boundaries of the field (second row), while 
incorrectly recognizing the activity as ``tennis swinging''. To see whether we can
potentially correct the model's mistake by forcing it to look at the relevant locations,
we took a trained model and initialized the location softmax weights to uniform random
numbers between the minimum and maximum in the original model. The model's glimpse in
this case is shown in the third row of \Figref{fig:blankagain}. We next 
optimized only the softmax weights, or the location variables, for this specific example
of ``soccer juggling'' to find the glimpse for which the model would predict it
correctly. All the other model parameters were kept fixed. Note that this only changes
the sequences of glimpses, or where the model attends to, and not the model itself. It
is interesting to see that in order to classify this video clip correctly, the glimpse
the model learns (the fourth row of \Figref{fig:blankagain}) tends to focus on the
soccer player's legs.

\section{Conclusion}\label{sec:conclusion}
\vspace{-0.05in}
In this paper we developed recurrent soft attention based models for action recognition
and analyzed where they focus their attention. Our proposed model tends to recognize important elements in video frames based
on the action that is being performed. We also showed that our model performs better
than baselines which do not use any attention mechanism. Soft attention models, though
impressive, are still computationally expensive since they still require all the
features to perform dynamic pooling. In the future, we plan to explore hard attention
models as well as hybrid soft and hard attention approaches which can reduce the
computational cost of our model, so that we can potentially scale to larger datasets like UCF-101 and the
Sports-1M dataset. These models can also be extended to the multi-resolution setting, in
which the attention mechanism could also choose to focus on the earlier convolutional
layers in order to attend to the lower-level features in the video frames.

{\small 
{\bf Acknowledgments}:
This work was supported by IARPA and Raytheon BBN Contract No. D11PC20071.
We would like to thank Nitish Srivastava for valuable discussions and  
Yukun Zhu for his assistance with the CNN packages.
}

\bibliography{iclr2016_workshop}

\begin{thebibliography}{35}
\providecommand{\natexlab}[1]{#1}
\providecommand{\url}[1]{\texttt{#1}}
\expandafter\ifx\csname urlstyle\endcsname\relax
  \providecommand{\doi}[1]{doi: #1}\else
  \providecommand{\doi}{doi: \begingroup \urlstyle{rm}\Url}\fi

\bibitem[Ba et~al.(2015{\natexlab{a}})Ba, Grosse, Salakhutdinov, and
  Frey]{Ba2015}
J.~Ba, R.~Grosse, R.~Salakhutdinov, and B.~Frey.
\newblock Learning wake-sleep recurrent attention models.
\newblock In \emph{NIPS}, 2015{\natexlab{a}}.

\bibitem[Ba et~al.(2015{\natexlab{b}})Ba, Mnih, and
  Kavukcuoglu]{DBLP:journals/corr/BaMK14}
J.~Ba, V.~Mnih, and K.~Kavukcuoglu.
\newblock Multiple object recognition with visual attention.
\newblock \emph{ICLR}, 2015{\natexlab{b}}.

\bibitem[Bahdanau et~al.(2015)Bahdanau, Cho, and
  Bengio]{DBLP:journals/corr/BahdanauCB14}
D.~Bahdanau, K.~Cho, and Y.~Bengio.
\newblock Neural machine translation by jointly learning to align and
  translate.
\newblock \emph{ICLR}, 2015.

\bibitem[Bastien et~al.(2012)Bastien, Lamblin, Pascanu, Bergstra, Goodfellow,
  Bergeron, Bouchard, Warde{-}Farley, and Bengio]{theano}
F.~Bastien, P.~Lamblin, R.~Pascanu, J.~Bergstra, I.~J. Goodfellow, A.~Bergeron,
  N.~Bouchard, D.~Warde{-}Farley, and Y.~Bengio.
\newblock Theano: new features and speed improvements.
\newblock \emph{CoRR}, abs/1211.5590, 2012.

\bibitem[Deng et~al.(2009)Deng, Dong, Socher, Li, Li, and
  Li]{DBLP:conf/cvpr/DengDSLL009}
J.~Deng, W.~Dong, R.~Socher, L.{-}J. Li, K.~Li, and F.{-}F. Li.
\newblock Imagenet: {A} large-scale hierarchical image database.
\newblock In \emph{CVPR}, 2009.

\bibitem[Donahue et~al.(2015)Donahue, Hendricks, Guadarrama, Rohrbach,
  Venugopalan, Saenko, and Darrell]{DBLP:journals/corr/DonahueHGRVSD14}
J.~Donahue, L.~A. Hendricks, S.~Guadarrama, M.~Rohrbach, S.~Venugopalan,
  K.~Saenko, and T.~Darrell.
\newblock Long-term recurrent convolutional networks for visual recognition and
  description.
\newblock In \emph{CVPR}, 2015.

\bibitem[Fernando et~al.(2015)Fernando, Gavves, Oramas, Ghodrati, and
  Tuytelaars]{Fernando2015a}
B.~Fernando, E.~Gavves, J.~Oramas, A.~Ghodrati, and T.~Tuytelaars.
\newblock Modeling video evolution for action recognition.
\newblock In \emph{CVPR}, 2015.

\bibitem[Graves et~al.(2013)Graves, Jaitly, and
  Mohamed]{DBLP:conf/asru/GravesJM13}
A.~Graves, N.~Jaitly, and A.{-}r. Mohamed.
\newblock Hybrid speech recognition with deep bidirectional {LSTM}.
\newblock In \emph{2013 {IEEE} Workshop on Automatic Speech Recognition and
  Understanding}, pp.\  273--278. {IEEE}, 2013.

\bibitem[Hochreiter \& Schmidhuber(1997)Hochreiter and
  Schmidhuber]{DBLP:journals/neco/HochreiterS97}
S.~Hochreiter and J.~Schmidhuber.
\newblock Long short-term memory.
\newblock \emph{Neural Computation}, 9\penalty0 (8):\penalty0 1735--1780, 1997.

\bibitem[Jaderberg et~al.(2015)Jaderberg, Simonyan, Zisserman, and
  Kavukcuoglu]{DBLP:journals/corr/JaderbergSZK15}
M.~Jaderberg, K.~Simonyan, A.~Zisserman, and K.~Kavukcuoglu.
\newblock Spatial transformer networks.
\newblock \emph{CoRR}, abs/1506.02025, 2015.

\bibitem[Jain et~al.(2015)Jain, v.~Gemert, and Snoek]{JainCVPR15}
M.~Jain, J.~C. v.~Gemert, and C.~G.~M. Snoek.
\newblock What do 15,000 object categories tell us about classifying and
  localizing actions?
\newblock In \emph{CVPR}, June 2015.

\bibitem[Karpathy et~al.(2014)Karpathy, Toderici, Shetty, Leung, Sukthankar,
  and Li]{DBLP:conf/cvpr/KarpathyTSLSF14}
A.~Karpathy, G.~Toderici, S.~Shetty, T.~Leung, R.~Sukthankar, and F.{-}F. Li.
\newblock Large-scale video classification with convolutional neural networks.
\newblock In \emph{CVPR}, 2014.

\bibitem[Kingma \& Ba(2015)Kingma and Ba]{DBLP:journals/corr/KingmaB14}
D.~P. Kingma and J.~Ba.
\newblock Adam: {A} method for stochastic optimization.
\newblock \emph{ICLR}, 2015.

\bibitem[Lan et~al.(2014)Lan, Lin, Li, Hauptmann, and
  Raj]{DBLP:journals/corr/LanLLHR14}
Z.{-}Z. Lan, M.~Lin, X.~Li, A.~G. Hauptmann, and B.~Raj.
\newblock Beyond gaussian pyramid: Multi-skip feature stacking for action
  recognition.
\newblock \emph{CoRR}, abs/1411.6660, 2014.

\bibitem[Mnih et~al.(2014)Mnih, Heess, Graves, and
  Kavukcuoglu]{DBLP:conf/nips/MnihHGK14}
V.~Mnih, N.~Heess, A.~Graves, and K.~Kavukcuoglu.
\newblock Recurrent models of visual attention.
\newblock In \emph{NIPS}, 2014.

\bibitem[Netzer et~al.(2011)Netzer, Wang, Coates, Bissacco, Wu, and Ng]{svhn}
Y.~Netzer, T.~Wang, A.~Coates, A.~Bissacco, B.~Wu, and A.~Y. Ng.
\newblock Reading digits in natural images with unsupervised feature learning.
\newblock In \emph{NIPS Workshop on Deep Learning and Unsupervised Feature
  Learning 2011}, 2011.

\bibitem[Ng et~al.(2015)Ng, Hausknecht, Vijayanarasimhan, Vinyals, Monga, and
  Toderici]{DBLP:journals/corr/NgHVVMT15}
J.~Y.{-}H. Ng, M.~J. Hausknecht, S.~Vijayanarasimhan, O.~Vinyals, R.~Monga, and
  G.~Toderici.
\newblock Beyond short snippets: Deep networks for video classification.
\newblock In \emph{CVPR}, 2015.

\bibitem[Peng et~al.(2014)Peng, Zou, Qiao, and Peng]{DBLP:conf/eccv/PengZQP14}
X.~Peng, C.~Zou, Y.~Qiao, and Q.~Peng.
\newblock Action recognition with stacked fisher vectors.
\newblock In \emph{{ECCV}}, volume 8693, pp.\  581--595. Springer, 2014.

\bibitem[Ren et~al.(2015)Ren, He, Girshick, Zhang, and
  Sun]{DBLP:journals/corr/RenHGZ015}
S.~Ren, K.~He, R.~B. Girshick, X.~Zhang, and J.~Sun.
\newblock Object detection networks on convolutional feature maps.
\newblock \emph{CoRR}, abs/1504.06066, 2015.

\bibitem[Rensink(2000)]{rensink2000dynamic}
R.~A. Rensink.
\newblock The dynamic representation of scenes.
\newblock \emph{Visual Cognition}, 7\penalty0 (1-3):\penalty0 17--42, 2000.

\bibitem[Simonyan \& Zisserman(2014)Simonyan and
  Zisserman]{DBLP:journals/corr/SimonyanZ14}
K.~Simonyan and A.~Zisserman.
\newblock Two-stream convolutional networks for action recognition in videos.
\newblock In \emph{NIPS}. 2014.

\bibitem[Srivastava et~al.(2014)Srivastava, Hinton, Krizhevsky, Sutskever, and
  Salakhutdinov]{DBLP:journals/jmlr/SrivastavaHKSS14}
N.~Srivastava, G.~E. Hinton, A.~Krizhevsky, I.~Sutskever, and R.~Salakhutdinov.
\newblock Dropout: a simple way to prevent neural networks from overfitting.
\newblock \emph{JMLR}, 15\penalty0 (1):\penalty0 1929--1958, 2014.

\bibitem[Srivastava et~al.(2015)Srivastava, Mansimov, and
  Salakhutdinov]{DBLP:journals/corr/SrivastavaMS15}
N.~Srivastava, E.~Mansimov, and R.~Salakhutdinov.
\newblock Unsupervised learning of video representations using {LSTM}s.
\newblock \emph{ICML}, 2015.

\bibitem[Sun et~al.(2014)Sun, Jia, Chan, Fang, Wang, and
  Yan]{DBLP:conf/cvpr/SunJCFWY14}
L.~Sun, K.~Jia, T.{-}H. Chan, Y.~Fang, G.~Wang, and S.~Yan.
\newblock {DL-SFA:} deeply-learned slow feature analysis for action
  recognition.
\newblock In \emph{CVPR}, 2014.

\bibitem[Sutskever et~al.(2014)Sutskever, Vinyals, and Le]{Ilya}
I.~Sutskever, O.~Vinyals, and Q.~V.~V. Le.
\newblock Sequence to sequence learning with neural networks.
\newblock In \emph{NIPS}. 2014.

\bibitem[Szegedy et~al.(2015)Szegedy, Liu, Jia, Sermanet, Reed, Anguelov,
  Erhan, Vanhoucke, and Rabinovich]{DBLP:journals/corr/SzegedyLJSRAEVR14}
C.~Szegedy, W.~Liu, Y.~Jia, P.~Sermanet, S.~Reed, D.~Anguelov, D.~Erhan,
  V.~Vanhoucke, and A.~Rabinovich.
\newblock Going deeper with convolutions.
\newblock In \emph{CVPR}, 2015.

\bibitem[Venugopalan et~al.(2014)Venugopalan, Xu, Donahue, Rohrbach, Mooney,
  and Saenko]{DBLP:journals/corr/VenugopalanXDRMS14}
S.~Venugopalan, H.~Xu, J.~Donahue, M.~Rohrbach, R.~J. Mooney, and K.~Saenko.
\newblock Translating videos to natural language using deep recurrent neural
  networks.
\newblock \emph{CoRR}, abs/1412.4729, 2014.

\bibitem[Vinyals et~al.(2015)Vinyals, Toshev, Bengio, and
  Erhan]{DBLP:journals/corr/VinyalsTBE14}
O.~Vinyals, A.~Toshev, S.~Bengio, and D.~Erhan.
\newblock Show and tell: {A} neural image caption generator.
\newblock In \emph{CVPR}, 2015.

\bibitem[Williams(1992)]{DBLP:journals/ml/Williams92}
R.~J. Williams.
\newblock Simple statistical gradient-following algorithms for connectionist
  reinforcement learning.
\newblock \emph{Machine Learning}, 8:\penalty0 229--256, 1992.

\bibitem[Wu et~al.(2015)Wu, Yan, Shan, Dang, and
  Sun]{DBLP:journals/corr/WuYSDS15}
R.~Wu, S.~Yan, Y.~Shan, Q.~Dang, and G.~Sun.
\newblock Deep image: Scaling up image recognition.
\newblock \emph{CoRR}, abs/1501.02876, 2015.

\bibitem[Xu et~al.(2015)Xu, Ba, Kiros, Cho, Courville, Salakhutdinov, Zemel,
  and Bengio]{DBLP:journals/corr/XuBKCCSZB15}
K.~Xu, J.~Ba, R.~Kiros, K.~Cho, A.~C. Courville, R.~Salakhutdinov, R.~S. Zemel,
  and Y.~Bengio.
\newblock Show, attend and tell: Neural image caption generation with visual
  attention.
\newblock \emph{ICML}, 2015.

\bibitem[Yao et~al.(2015)Yao, Torabi, Cho, Ballas, Pal, Larochelle, and
  Courville]{DBLP:journals/corr/YaoTCBPLC15}
L.~Yao, A.~Torabi, K.~Cho, N.~Ballas, C.~Pal, H.~Larochelle, and A.~Courville.
\newblock Describing videos by exploiting temporal structure.
\newblock \emph{CoRR}, abs/1502.08029, 2015.

\bibitem[Yeung et~al.(2015)Yeung, Russakovsky, Jin, Andriluka, Mori, and
  Li]{DBLP:journals/corr/YeungRJAML15}
S.~Yeung, O.~Russakovsky, N.~Jin, M.~Andriluka, G.~Mori, and F.{-}F. Li.
\newblock Every moment counts: Dense detailed labeling of actions in complex
  videos.
\newblock \emph{CoRR}, abs/1507.05738, 2015.

\bibitem[Zaremba et~al.(2014)Zaremba, Sutskever, and
  Vinyals]{DBLP:journals/corr/ZarembaSV14}
W.~Zaremba, I.~Sutskever, and O.~Vinyals.
\newblock Recurrent neural network regularization.
\newblock \emph{CoRR}, abs/1409.2329, 2014.

\bibitem[Zha et~al.(2015)Zha, Luisier, Andrews, Srivastava, and
  Salakhutdinov]{arXiv_video}
S.~Zha, F.~Luisier, W.~Andrews, N.~Srivastava, and R.~Salakhutdinov.
\newblock Exploiting image-trained {CNN} architectures for unconstrained video
  classification.
\newblock \emph{CoRR}, abs/1503.04144, 2015.

\end{thebibliography}
\bibliographystyle{iclr2016_workshop}

\newpage
\appendix
\section{Additional examples}
\label{app:examples}
\begin{figure}[!hb]
\begin{minipage}{\textwidth}
\hspace{0.1in}
\begin{subfigure}{0.45\textwidth}
\includegraphics[width=\linewidth]{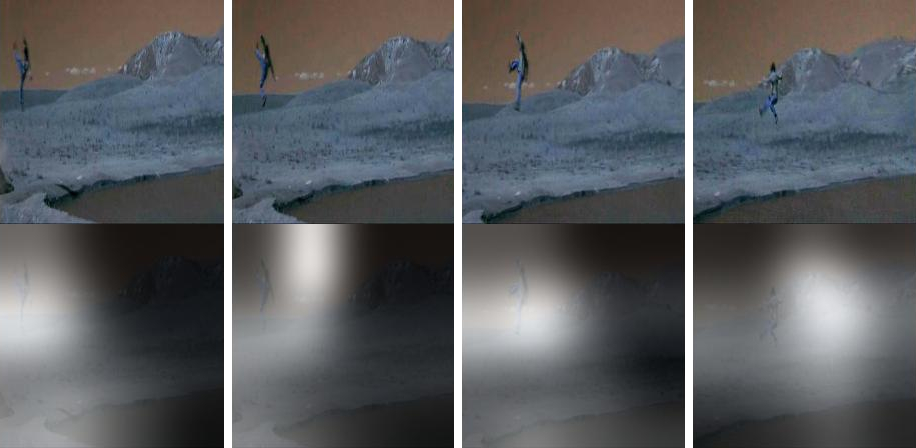}
\caption{``dive''}
\vspace{0.2in}
\label{subfig-mid-a1}
\end{subfigure}
\begin{subfigure}{0.10\textwidth}
\end{subfigure}
\begin{subfigure}{0.45\textwidth}
\includegraphics[width=\linewidth]{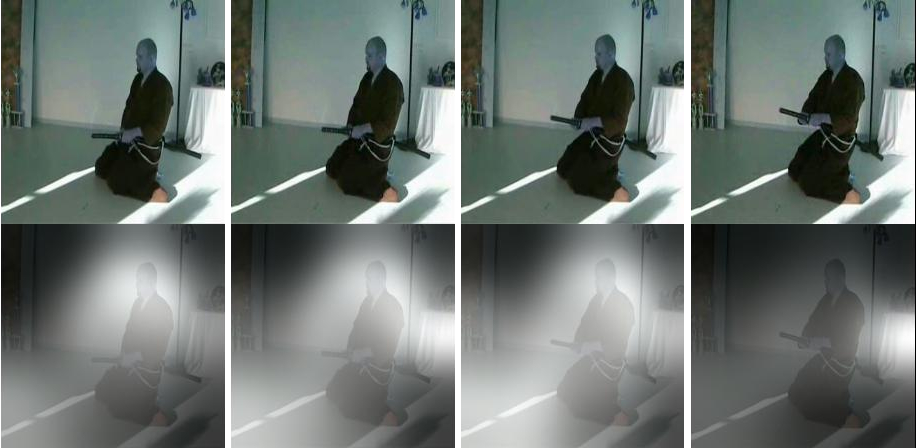}
\caption{``draw\_sword''}
\vspace{0.2in}
\label{subfig-mid-a2}
\end{subfigure}
\end{minipage}
\begin{minipage}{\textwidth}
\hspace{0.1in}
\begin{subfigure}{0.45\textwidth}
\includegraphics[width=\linewidth]{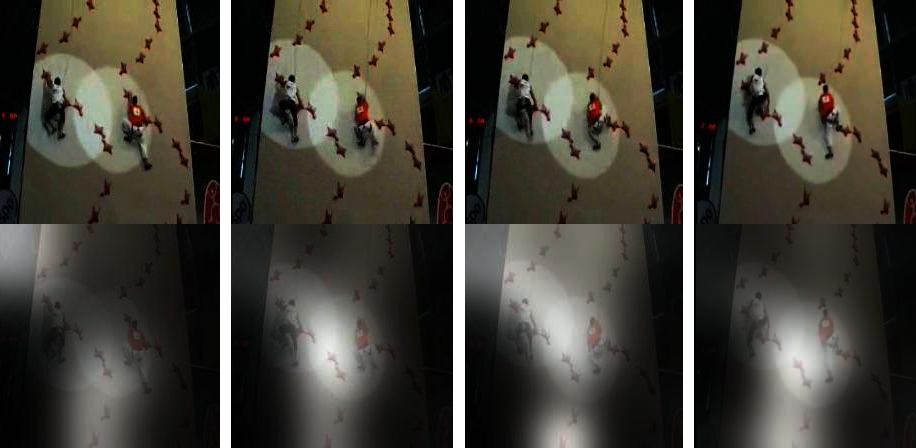}
\caption{``climb''}
\vspace{0.2in}
\label{subfig-mid-a3}
\end{subfigure}
\begin{subfigure}{0.10\textwidth}
\end{subfigure}
\begin{subfigure}{0.45\textwidth}
\includegraphics[width=\linewidth]{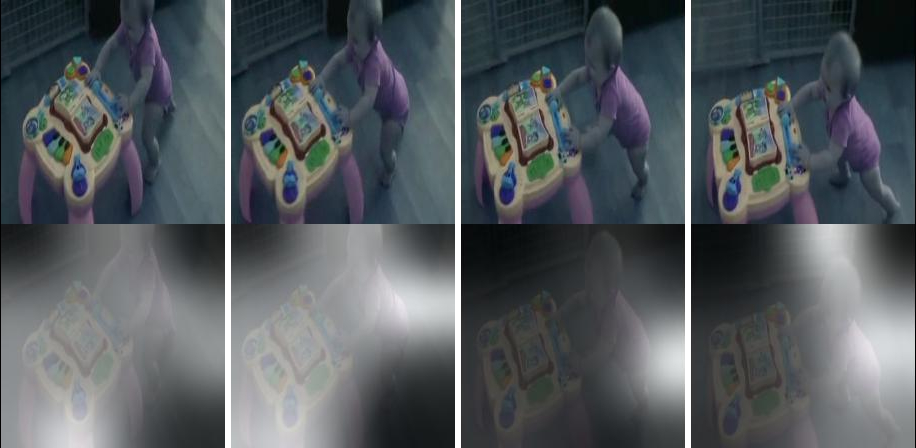}
\caption{``push''}
\vspace{0.2in}
\label{subfig-mid-a4}
\end{subfigure}
\end{minipage}
\begin{minipage}{\textwidth}
\hspace{0.1in}
\begin{subfigure}{0.45\textwidth}
\includegraphics[width=\linewidth]{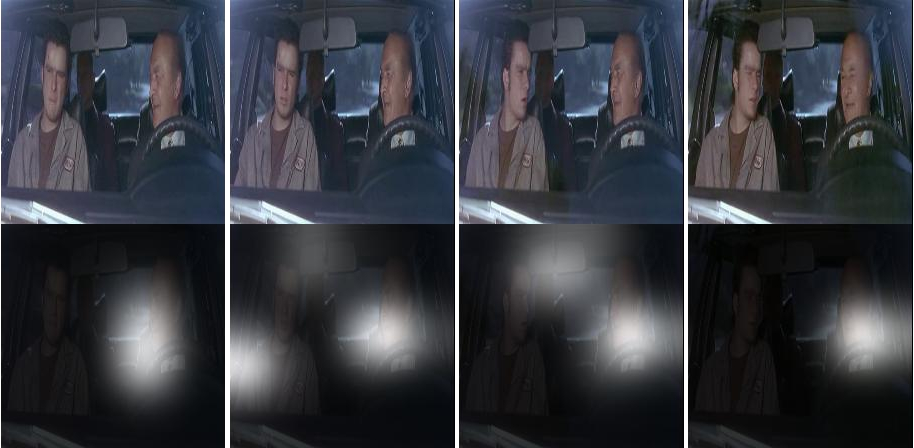}
\caption{``DriveCar''}
\label{subfig-mid-a5}
\end{subfigure}
\begin{subfigure}{0.10\textwidth}
\end{subfigure}
\begin{subfigure}{0.45\textwidth}
\includegraphics[width=\linewidth]{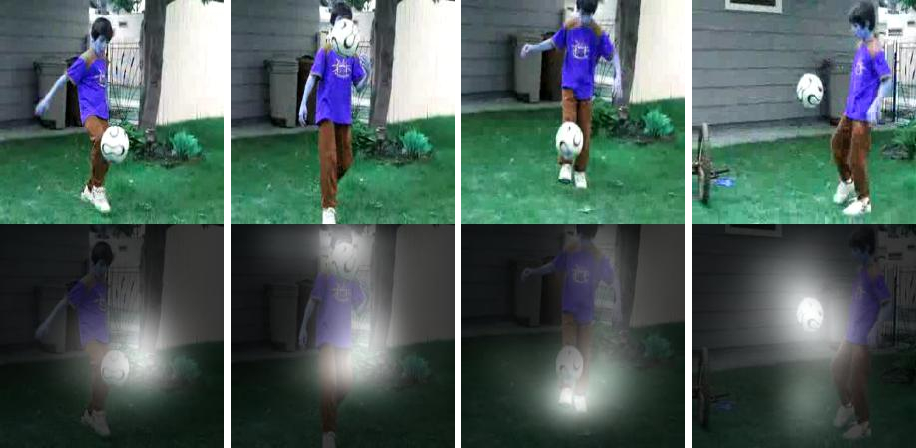}
\caption{``soccer juggling''}
\label{subfig-mid-a6}
\end{subfigure}
\end{minipage}
\caption{\small Correctly classified video frames showing attention over time: The white regions are where the model is
looking and the brightness indicates the strength of focus. The model learns to look at relevant parts.}
\label{fig:midfig-appendix}
\end{figure}

\begin{figure}[!hb]
\begin{minipage}{\textwidth}
\hspace{0.1in}
\begin{subfigure}{0.45\textwidth}
\includegraphics[width=\linewidth]{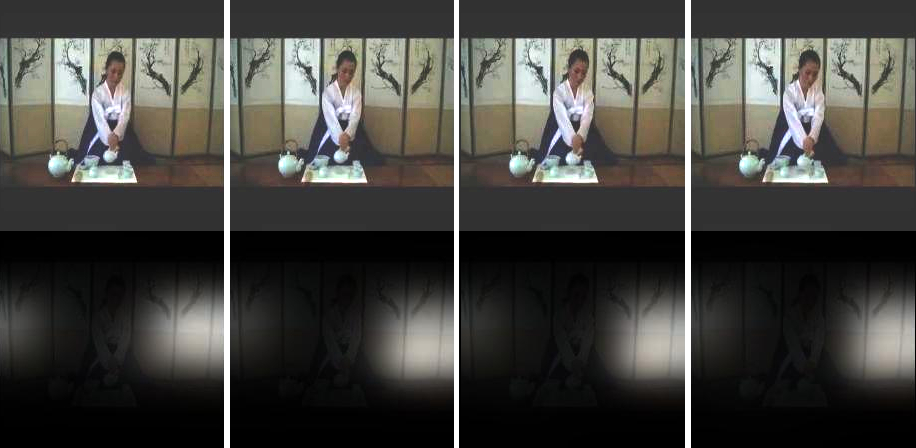}
\caption{``pour'' misclassified as ``push''}
\label{subfig-mid-a1}
\end{subfigure}
\begin{subfigure}{0.10\textwidth}
\end{subfigure}
\begin{subfigure}{0.45\textwidth}
\includegraphics[width=\linewidth]{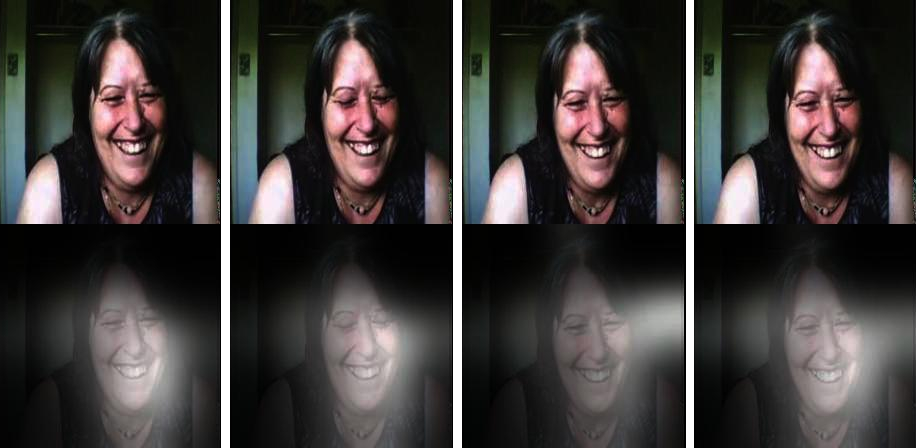}
\caption{``laugh'' misclassified as ``smile''}
\label{subfig-mid-a2}
\end{subfigure}
\end{minipage}
\caption{\small Incorrectly classified video frames showing attention over time: The white regions are where the model is looking and the brightness indicates the strength of focus.}
\label{fig:midfig-incorrect-appendix}
\end{figure}

\vspace{0.2in}
We present some more correctly classified examples from UCF-11, HMDB-51 and Hollywood2 in \Figref{fig:midfig-appendix} and incorrectly classified examples in \Figref{fig:midfig-incorrect-appendix}.

\end{document}